\documentclass{article}
\usepackage[preprint]{icml2026}
\usepackage[utf8]{inputenc}
\usepackage[T1]{fontenc}
\usepackage{booktabs}
\usepackage{graphicx}
\usepackage{amsmath,amssymb}
\usepackage{xcolor}
\usepackage{natbib}
\usepackage[hidelinks]{hyperref}
\usepackage{multirow}
\usepackage{array}
\usepackage{seqsplit}

\newcolumntype{L}[1]{>{\raggedright\arraybackslash}p{#1}}
\graphicspath{{figures/}}
\newcommand{\arXivAuthor}{Dongwu Qian}
\newcommand{\arXivAffiliation}{Independent Researcher}
\newcommand{\arXivEmail}{qiandongwu123@qq.com}

\icmltitlerunning{Counting Events, Not Seeing Them}

\begin{document}

\twocolumn[
\icmltitle{Counting Events, Not Seeing Them: Diagnosing and Repairing\\
the Readout Failure in Video LLMs}

\begin{icmlauthorlist}
\icmlauthor{\arXivAuthor}{aff}
\end{icmlauthorlist}
\icmlaffiliation{aff}{\arXivAffiliation}
\icmlcorrespondingauthor{\arXivAuthor}{\arXivEmail}
\icmlkeywords{video LLMs, temporal perception, object counting, linear probing, evaluation audit}

\vskip 0.3in
]
\hypersetup{pdftitle={Counting Events, Not Seeing Them: Diagnosing and Repairing the Readout Failure in Video LLMs},pdfauthor={\arXivAuthor}}
\printAffiliationsAndNotice{}

\begin{abstract}
Video LLMs fail conspicuously at counting how many times an object crosses the screen, a failure attributed to missing object permanence. Using construction-grounded synthetic testbeds with independently validated labels, we show this attribution is incomplete: the count signal is \emph{present but unread}---linear probes decode per-window presence at 93\% while global count decoding collapses to a duration confound. Concurrent work reports the same representation--output dissociation for repetition counting; we show that in exit-and-reappear event counting it is repairable: two readout-directed repairs (episode-localization supervision and a discrete event-symbol side channel) raise same-object counting from 46.7\% to 65.0\% (oracle 66.7\%) on a chance-33\% benchmark across 7B--30B systems. Transfer to an independent factory costs nothing, and a measurement audit exposes five data-integrity failures that had obscured the boundary. A real-video audit (REALCOUNT, $n=18$, blind-audited) shows the boundary persists---best 55.6\% at chance 33.3\%---and symbol transfer fails zero-shot. A 1-epoch in-domain adaptation bench is negative (six levers at 28--50\%), but a 3-epoch schedule shows this is budget-limited for the symbol channel: with oracle timelines the reader reaches 94.4\% sequence-disjoint (and 100\% strict on the full training set), while the deployed 2-epoch head reaches 61.1\% in all three seeds (42--44)---no better than a shuffled-symbol control (61.1\% in seed 42, 66.7\% in seed 43) and one clip above the unadapted reader (55.6\%). The residual gap is detector evidence, not reader capacity: the same reader reaches 72.2--88.9\% on GT timelines across those seeds. The readout failure is repairable with informative timelines, but the deployed head remains the bottleneck; a frozen-feature boundary probe finds the signal weak (AUC 0.62 over a 0.51 permutation control) and non-transferring across clips (0.77 within-clip vs.\ 0.62 across windows).
\end{abstract}

\section{Introduction}
\label{sec:intro}

Ask a video LLM how many times the red square crossed the screen, and it will fail---not occasionally but systematically, across model scales and training regimes. This ``same-object counting'' failure has survived every axis it has been tested on: 7B and 30B models, zero-shot and fine-tuned, dense and mixture-of-experts, attention-LoRA and expert-LoRA (Sec.~\ref{sec:diag}). Because \emph{heterogeneous-object} counting is near-perfect for the same models, the deficit has been characterized as missing \textbf{object permanence}---a representational gap in how identity is maintained across time \citep{baillargeon1987object}.

This paper argues the deficit is elsewhere, and that it is repairable. Our diagnosis pipeline---error-structure analysis against trivial-baseline controls, linear probing of the model's own token stream, and temporal-granularity measurements---locates the failure in the \textbf{readout}: the information needed to count is present in the model's visual token stream (a linear probe recovers per-window object presence at 93\%), but the language model cannot aggregate it into a count. Counting, for a video LLM, is not perception; it is symbol manipulation over evidence it can already see.

Concurrent work reaches the same conclusion for repetition counting \citep{khoshnoodi2026reprobe}. We extend it in two directions: the regime is exit-and-reappear event counting on real video rather than periodic action repetition, and the dissociation is not merely diagnosed but repaired.

This reframing has a practical consequence: if the bottleneck is readout, the fix is not better visual features but \emph{structured readout pathways}. We evaluate two, both trained on construction-grounded data with episode-level labels that our factories emit for free:
\begin{enumerate}
\item \textbf{Episode-localization supervision} (output-side): half the MCQ budget is replaced by a task asking the model to state the start and end time of every crossing. The model learns to articulate the event structure it is asked to count (episode recall 84\% at IoU$\ge$0.5).
\item \textbf{An event-symbol side channel} (input-side): a lightweight learned detector (a 3-class head over frozen connector features with edge-band spatial pooling) emits one symbol per half-second---\texttt{\#} present, \texttt{.} absent, \texttt{|} transition---injected into the prompt at a cost of one token per slot. Counting becomes counting \emph{symbols}, the regime in which the same models are near-perfect.
\end{enumerate}
Both interventions improve over identically-budgeted clean-data training by +10.0 to +15.0 points (paired bootstrap, $P\le.01$--$.05$), reaching 65.0\% deployable / 66.7\% oracle on a chance-33\% benchmark where every prior configuration sat at 43--52\%.

Along the way we audit the measurement apparatus itself, and report five data-integrity findings that changed the reported numbers: an evaluation set fully contained in training data (producing ``impossible'' centisecond precision that was memorization), a timestamp-precision artifact, a label error in a reused motion formula that made ``continuous crossing'' videos contain a single crossing, a training-window episode-truncation bug that wrote 64 episode ends to the clip end, and a training-prompt class mismatch in 29 of 145 real-video windows. We distill these into auditable rules (Appendix~\ref{app:integrity}).

\textbf{Contributions.}
\begin{itemize}
\item A controlled re-characterization of the same-object counting boundary that independently confirms and extends the representation--output dissociation reported for repetition counting \citep{khoshnoodi2026reprobe}: prior-matched evaluation, confound-controlled probing, and a granularity analysis together relocate the deficit from representation to readout (Sec.~\ref{sec:diag}).
\item Two readout-directed repairs---localization supervision and a deployable event-symbol side channel---with a full-stack combination, noise-matched training, and an edge-band detector head; each component evaluated with paired bootstrap statistics (Secs.~\ref{sec:inter}--\ref{sec:results}).
\item A measurement-integrity audit of the evaluation infrastructure itself, with three documented failure modes and their detection rules (Appendix~\ref{app:integrity}).
\item All data factories emit construction-grounded episode labels validated by independent measurement (93--99.7\% agreement), and every reported number is reproducible from archived records.
\item A real-video audit (REALCOUNT, Sec.~\ref{sec:real}): an exit-and-reappear counting set built from long-term tracking benchmarks, where every one of the final 18 items survived a blind multi-pass frame-level audit (23\% yield from 78 candidates)---showing the boundary persists on natural footage and that neither the probe readability nor the symbol channel transfers zero-shot.
\item A six-lever in-domain adaptation bench that, at a 1-epoch budget, fails to exceed the unadapted baseline---and a 3-epoch schedule showing that the symbol-channel lever reverses this once trained (94.4\% with sequence-disjoint oracle timelines, 100\% strict on the full set), while the deployed 2-epoch head reaches only 61.1\% and does not beat a shuffled-symbol control (61.1\% in seed 42, 66.7\% in seed 43)---plus a boundary probe that localizes the residual real-video gap to the frozen evidence rather than the adapter (Sec.~\ref{sec:ceiling}).
\end{itemize}

\section{Related Work}
\label{sec:related}

\textbf{Video LLMs and temporal perception.} Dominant video-LLM architectures sample frames uniformly and encode them per-frame \citep{xu2025qwen25omni}; documented failures include motion blindness and temporal-grounding errors \citep{liu2024tempcompass,fu2024videomme}. Fine-grained temporal tasks (counting repeated events, ordering, speed) remain weak even at frontier scale---our Sec.~\ref{sec:diag} confirms this across a 7B$\to$30B sweep---motivating diagnosis rather than scale.

\textbf{Probing and diagnosis.} Linear probing to localize information within frozen representations is standard \citep{alain2016probe}. We extend it with two controls specific to temporal benchmarks: a \emph{duration confound} baseline (clip length predicts event count in most generated data) and a permutation control, without which probe results overstate what tokens carry.

\textbf{Object permanence and event segmentation.} Cognitive work distinguishes representing continuity from enumerating discrete events \citep{baillargeon1987object,zacks2007event}. Our results operationalize this distinction in video LLMs: presence (continuity) is linearly readable; enumeration is not.

\textbf{Concurrent diagnosis of counting failures.} A 2026 wave of benchmarks measures video counting from complementary angles: repetition counting at controlled counts \citep{khoshnoodi2026reprobe,pushupbench2026}, trace-grounded event bookkeeping \citep{lowfrequencytrap2026}, momentary events under dense sampling \citep{momentvideo2026}, streaming state maintenance \citep{svcbench2026}, long-video enumeration \citep{ecbench2026}, numerical reasoning \citep{vidnum2026}, object counting under controlled cardinality \citep{countvideobench2026}, and object-centric temporal consistency including disappearance and reappearance \citep{tocbench2026}. Two findings recur: accuracy collapses at higher counts even when frames cover the evidence, and count information is often linearly decodable while the output is wrong \citep{khoshnoodi2026reprobe,lowfrequencytrap2026,momentvideo2026}. These works diagnose; they do not repair. We ask whether the dissociation is repairable in a real-video exit-and-reappear regime.

\textbf{Data contamination and measurement validity.} Benchmark leakage into training data is an active concern \citep{xu2024contamination}. Appendix~\ref{app:integrity} contributes five concrete failure modes from a single project's audit---set-inclusion leakage, memorization masquerading as precision, formula-reuse label errors, training-window episode truncation, and a training-prompt class mismatch---with cheap detection rules.

\textbf{Prompt-side structured evidence.} Feeding text extracted from the video into the LLM context is established for long-video QA: caption-style pipelines summarize frames or clips into textual timelines consumed by the LLM \citep{zhang2023llovi,wang2024videoagent}, and temporal-grounding models integrate timestamp or event tokens into the token stream itself \citep{qian2024momentor,ren2023timechat,huang2023vtimeLLM,wang2024groundedvideollm,guo2024trace}; tool-use frameworks render detector and tracker outputs as text for the solver \citep{suris2023vipergpt,gao2023pal,schick2023toolformer}.

\textbf{Specialist side channels and structured event evidence.} Repetition counting on real video has been studied outside LLMs \citep{dwibedi2020counting} and recently as a VLM capability benchmark \citep{khoshnoodi2026reprobe,sinha2024everyshot}. Closest to our channel, detector-grounded structured prompts improve image counting \citep{groundcount2026}, a specialist's scalar count helps when delivered as a discrete symbol rather than a density map \citep{crowdcue2026}, and training-free visual markers or object-state tokens improve temporal grounding and persistence \citep{markit2026,slotnarrative2026}. These deliver boxes, scalar counts, visual overlays, or object states; our channel delivers a per-slot three-state event timeline, and we test its causal role with no-symbol, shuffled-symbol, and sequence-disjoint controls. Against these neighbors, our event-symbol channel is deliberately minimal---one 3-class symbol per fixed half-second slot (one token, not a caption)---produced by a lightweight learned detector whose noise model is matched to the reader at training time (Sec.~\ref{sec:arm2}), and targeted at a diagnosed readout failure (same-object counting) rather than generic captioning, grounding, or retrieval. To our knowledge, no prior work pairs a per-slot event-presence symbol channel with a matched deployable detector and causal controls on real exit-and-reappear counting.

\section{Experimental Setup}
\label{sec:setup}

\textbf{Models.} The intervention platform is Qwen2.5-Omni-7B (Thinker) \citep{xu2025qwen25omni} with a unified-token-stream architecture: a frozen VideoMAE-based fast-flow connector \citep{tong2022videomae} injects 196-token virtual rows into the video token grid, and the top 4 LLM layers are unfrozen ($\sim$0.93B trainable + 11.6M connector). All intervention arms share an identical lineage (initialized from the same predecessor checkpoint), identical non-count data mixtures, and matched budgets: 2,159 steps, except arm2/arm3, which use 2,939 steps to cover the extended nogap corpus (Sec.~\ref{sec:arm2}). Baselines additionally include Qwen3-Omni-30B-A3B \citep{xu2025qwen3omni} (4-bit, zero-shot and with a published expert-LoRA adapter).

\textbf{Data factories.} All counting data are synthetic with construction-grounded ground truth. \texttt{count\_v4}: red squares crossing real-video backgrounds; \emph{gap} variant ($k$ independent crossings separated by off-screen intervals, 0.4--0.7 s) and \emph{nogap} variant (continuous mod-wraparound crossings); $k\in\{1,2,3\}$ exactly balanced across six (variant $\times$ $k$) evaluation groups of 10 ($n=60$, chance 33.3\%). \texttt{v1x}: an independently written factory (different formula instantiation, prompts, and backgrounds; $n=90$) used \emph{only} for transfer evaluation. Episode labels (per-crossing $[start,end]$) are emitted by construction and verified by independent measurement---a shape-criterion detector (99.7\% agreement on held-out gap clips) and a centroid-wraparound counter (nogap).

\textbf{Protocol.} MCQ answers are parsed with a fixed letter grammar; every arm reports the constant-answer baseline alongside; paired bootstrap ($B=10{,}000$) over per-sample outcomes for all deltas. Reported $p$-values are uncorrected (no family-wise adjustment); the main contrasts were pre-specified in the archived run logs.

\section{Diagnosis: Present but Unread}
\label{sec:diag}

\subsection{The boundary, re-measured honestly}

\begin{table}[t]
\centering
\footnotesize
\caption{The historical ``46.7\% plateau'', re-measured on balanced v4.}
\begin{tabular}{@{}L{0.68\columnwidth}L{0.26\columnwidth}@{}}
\toprule
System & v4 count (chance 33.3\%) \\
\midrule
7B + transition-trained checkpoint (prior best) & 46.7 [33.3, 60.0] \\
7B same-mix, no-fast-flow retrain & 48.3 \\
30B zero-shot & 43.3 \\
30B + expert-LoRA (published recipe) & 33.3 (constant) \\
\bottomrule
\end{tabular}
\label{tab:boundary}
\end{table}

Two observations dismantle the naive reading of the historical ``46.7\% plateau''. First, it \emph{underperforms} the constant-most-frequent-class baseline of 60\%---the model was not partially counting; its predictions were prior-shaped plus a systematic over-segmentation bias (single crossings answered as 2 in 11/18 cases; count-3 essentially never produced). Second, the label prior itself was confounded: training data were count-2-heavy, evaluation count-1-heavy. Every arm in Table~\ref{tab:boundary} sits at 1.0--1.5$\times$ chance (Fig.~\ref{fig:diagnosis}a,b).

\subsection{Probing: local presence yes, global count no}

\begin{figure}[t]
\centering
\includegraphics[width=\columnwidth]{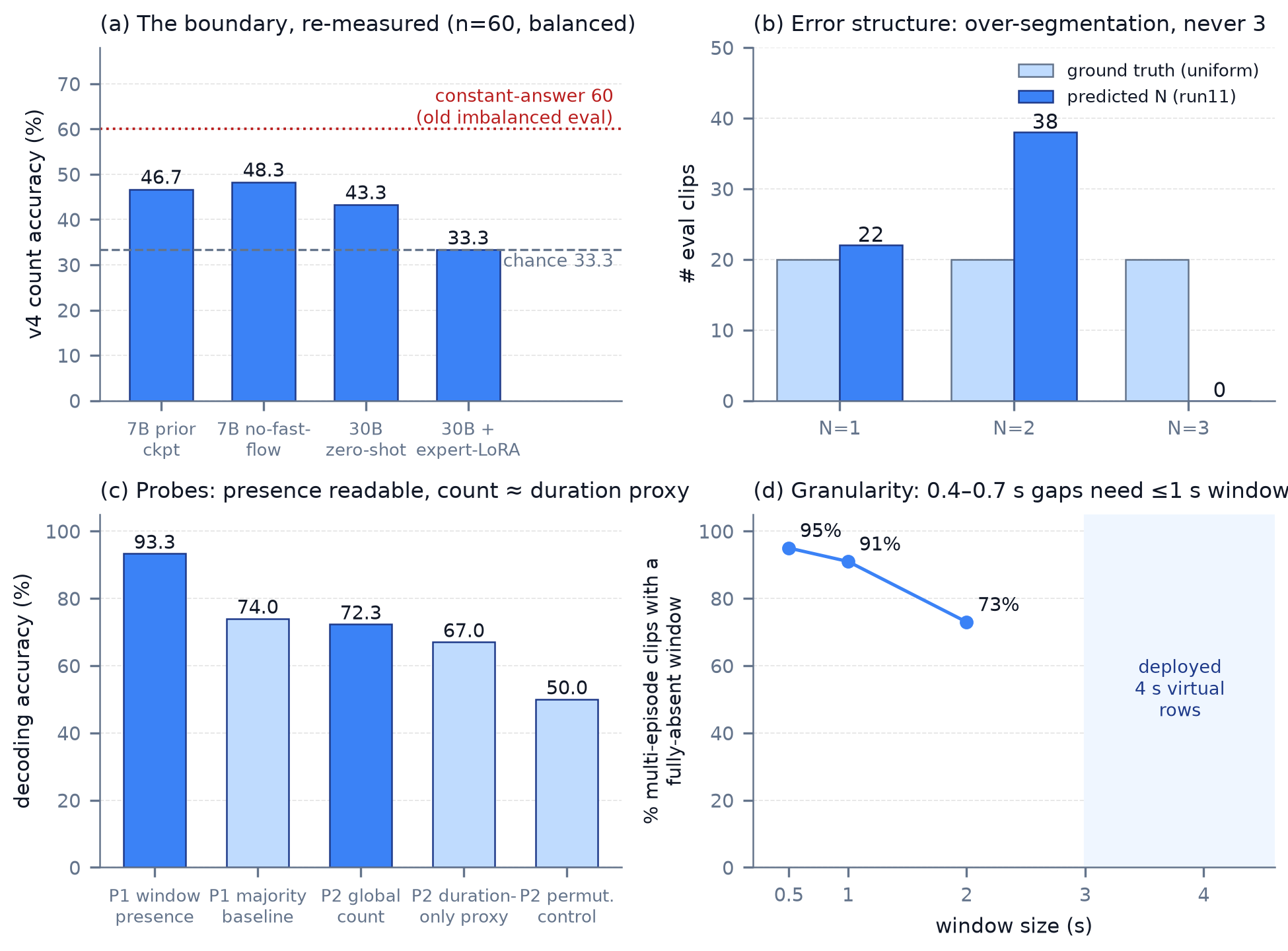}
\caption{\textbf{Diagnosis panel.} (a) The boundary re-measured on balanced v4 ($n=60$): every system sits at 1.0--1.5$\times$ chance and below the old eval's constant-answer baseline of 60. (b) Error structure of the prior checkpoint on v4: predicted counts never include 3 (0/60); $k=3$ clips are answered ``2'' in 18/20 cases. (c) Linear probes on the connector's token stream: local presence decodes at 93.3\% while global count decoding (72.3\%) barely exceeds a duration-only proxy (67\%); permutation control 50\%. (d) Granularity: the 0.4--0.7 s gaps between episodes are resolved only at $\le$1 s windows; the deployed 4 s virtual rows cannot separate episodes.}
\label{fig:diagnosis}
\end{figure}

With the connector's per-window token stream (what the LLM actually consumes as motion evidence):
\begin{itemize}
\item \textbf{P1 (window presence)}: 5-fold CV logistic regression decodes ``object visible in this 2 s window'' at \textbf{93.3\%} (majority baseline 74\%; absent-class recall 0.997).
\item \textbf{P2 (global count)}: pooled-token decoding reaches 72.3\%---but a \emph{duration-only} logistic baseline reaches 67\%, and a permutation control sits at 50\%. The apparent count signal is substantially a duration proxy; episode structure specifically is not linearly readable.
\end{itemize}
\textbf{P3 (granularity)}: the 0.4--0.7 s off-screen intervals that separate episodes are resolvable only at $\le$1 s windows (91\% of multi-episode clips produce a fully-absent window at 1 s; 95\% at 0.5 s; 73\% at 2 s). The 4 s virtual-row granularity of the deployed architecture is thus \emph{physically incapable} of episode separation---a design constraint, not a learned failure (Fig.~\ref{fig:diagnosis}c,d).

\subsection{The measurement was part of the problem}

Two label-integrity findings (details in Appendix~\ref{app:integrity}): the historical hard-cut evaluation set was fully contained in training data (its ``centisecond timestamp precision'' was memorization---provably, since the precision exceeded every input channel's physical resolution and survived time-reversed inputs); and a reused crossing formula was linear rather than periodic, silently converting $\sim$465 ``continuous crossing'' samples into single-crossing videos with multi-crossing labels. After repair and $k$-balancing, the boundary stands (Sec.~\ref{sec:diag})---but only now is it \emph{measured}.

\textbf{Diagnosis.} Perception: intact. Local representation: present. Global aggregation: absent. The deficit is a readout failure.

\section{Interventions}
\label{sec:inter}

All arms: identical lineage, budget, and non-count mixture; only the intervention differs.

\subsection{A-base (control): clean, balanced data}

Replacing the corrupted count data with \texttt{count\_v4} (true periodic formula, balanced $k$, episode labels) while holding everything else fixed: \textbf{51.7\%}. Clean data alone repairs the $k=1$ over-segmentation pathology (8/10 correct) but leaves $k\ge2$ near chance---the readout deficit persists through data repair (+5.0, CI [$-8.3$, $+18.3$] vs prior checkpoint).

\subsection{arm0: episode-localization supervision (output-side)}

Half the MCQ budget is exchanged for localization queries (``state the [start, end] interval of every crossing'', answers in JSON). \textbf{61.7\%}, with genuinely learned localization (episode recall \textbf{84.4\%}, precision 69.7\% at IoU$\ge$0.5; the untrained control produces out-of-range hallucinated intervals at 50\%/23.6\%). Paired $\Delta$ vs A-base: \textbf{+0.100 [+0.000, +0.217]}, $P(\Delta\le0)=0.047$. Notably this arm trains on \emph{half} the MCQ examples yet outperforms---supervision structure beats repetition.

\subsection{arm1: event-symbol side channel (input-side)}
\label{sec:arm1}

\begin{figure}[t]
\centering
\includegraphics[width=\columnwidth]{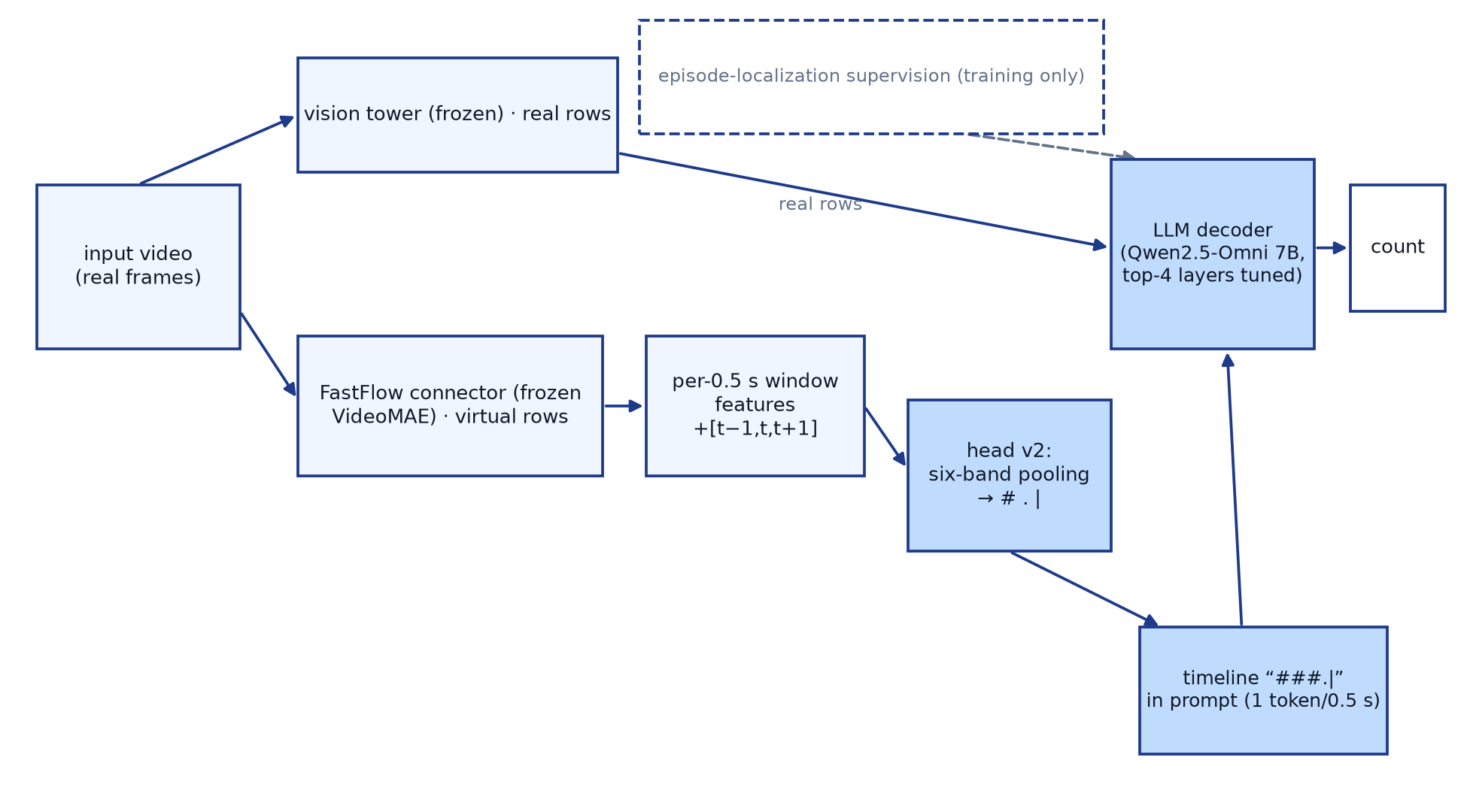}
\caption{\textbf{Event-symbol side channel.} Real rows flow through the frozen vision tower; the frozen fast-flow connector supplies virtual rows (motion evidence) whose per-0.5 s window features feed the independent head v2 (six-band spatial pooling over raw VideoMAE features). The resulting symbol timeline (one token per half-second slot) is appended to the prompt, converting same-object counting into symbol counting. Episode-localization supervision (dashed) acts on the output side during training only.}
\label{fig:arch}
\end{figure}

A presence head over the frozen connector's per-0.5 s window features (context-concatenated $[t-1,t,t+1]$) emits a three-class symbol per slot; timelines are appended to the prompt (\texttt{slot symbols: |\#\#\#||\#\#|}). Training uses ground-truth timelines with 10\% slot flips; inference uses head predictions. \textbf{63.3\%} (head pipeline) / 61.7\% (oracle timelines) / 55.0\% (symbols withheld). Three controls pin the mechanism: symbols injected into a symbol-untrained model do nothing (45.0\% $\approx$ baseline); random-feature ablations of the underlying channel eliminate the gain; and the model is robust to timeline noise (head $\approx$ oracle within noise). Figure~\ref{fig:arch} shows the channel architecture.

\subsection{arm2/arm3: full stack, noise matching, and the deployable head}
\label{sec:arm2}

Combining symbols + localization + extended nogap coverage (period spectrum 1.5--7.2 s): \textbf{oracle 66.7\%} (ceiling), head-v1 60.0\%. The gap between oracle and deployed timelines is closed not by prompt normalization (which slightly hurts) but by \textbf{noise-matched training} (arm3: train on head predictions; head $=$ oracle $=$ 60.0\%, confirming the distribution-matching hypothesis) and by \textbf{fixing the detector's inductive bias}: the ball-at-edge signal is spatial, but the connector's 196-token pooling mixes the spatial grid. Head v2 pools six spatial bands (left/right/top/bottom/center/global) directly from raw VideoMAE features---an independent branch that leaves the connector (and hence the LLM's trained alignment) untouched---lifting slot accuracy \textbf{64.7\% $\to$ 81.6\%} (per-class recall 77/82/84) and the deployed pipeline to \textbf{65.0\%}, 1.7 points under the oracle ceiling. The detector is base-model-agnostic by construction.

\section{Results}
\label{sec:results}

\subsection{Main table (v4, $n=60$, chance 33.3\%)}

\begin{table}[t]
\centering
\footnotesize
\caption{Main results on v4 with paired-bootstrap confidence intervals ($B=10{,}000$). All structural interventions are significant over A-base.}
\begin{tabular}{@{}L{0.39\columnwidth}L{0.27\columnwidth}L{0.25\columnwidth}@{}}
\toprule
Arm & Acc [95\% CI] & $\Delta$ vs A-base, $P$ \\
\midrule
Prior checkpoint (dirty data) & 46.7 [33.3, 60.0] & --- \\
30B zero-shot & 43.3 [31.7, 56.7] & --- \\
A-base (clean data) & 51.7 [38.3, 65.0] & --- \\
arm0 (+localization) & 61.7 [48.3, 73.3] & +0.100, .047 \\
arm1 (+symbols, head) & 63.3 [51.7, 75.0] & +0.117, .008 \\
arm2 (full stack, oracle) & 66.7 [55.0, 78.3] & +0.150, .010 \\
arm2 (full stack, head v2) & \textbf{65.0 [53.3, 76.7]} & +0.133, .008 \\
\bottomrule
\end{tabular}
\label{tab:main}
\end{table}

arm1 vs arm0: +0.017, $P=.444$---the two pathways are statistically indistinguishable; each is an independent witness to the readout hypothesis. Head v2 vs v1: +0.050, $P=.164$ (trend; the head's value is established at the slot level, $n=833$: 81.6\% vs 64.7\%). The full repair chain is summarized in Fig.~\ref{fig:waterfall}.

\subsection{Per-$k$ structure}

\begin{figure}[t]
\centering
\includegraphics[width=\columnwidth]{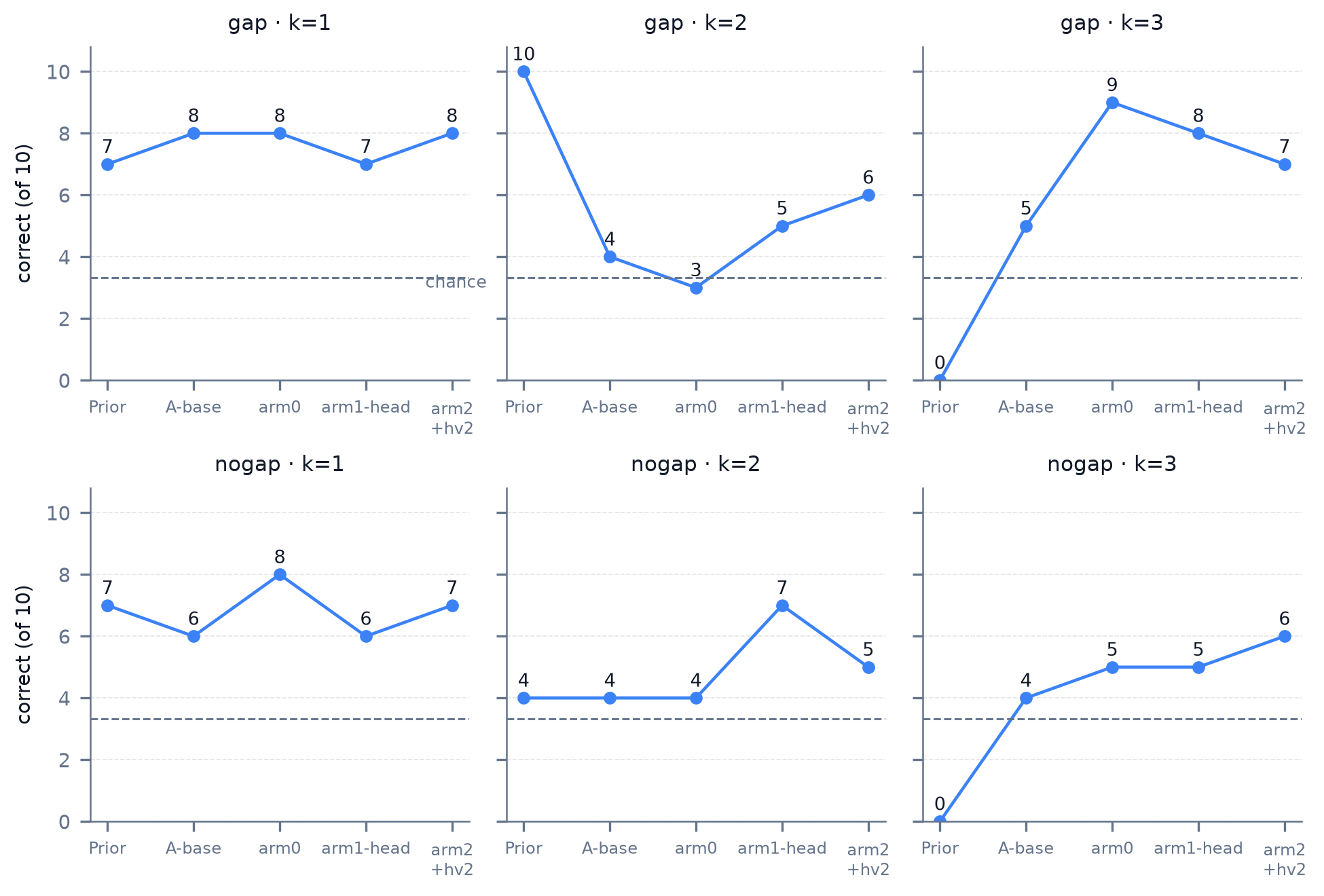}
\caption{\textbf{Per-cell accuracy across the repair chain} (each cell $n=10$, dashed line = chance 3.3/10). Clean data fixes gap $k=1$; the structural arms specifically lift gap $k=3$ (0/10 $\to$ 7--9/10); the $k=2/k=3$ boundary remains the shared residual, and nogap $k=2/k=3$ improve only under the full stack.}
\label{fig:perk}
\end{figure}

$k=1$ is solved by clean data alone (8--9/10 across arms). The structural interventions specifically repair \textbf{$k=3$} (gap: 5/10 $\to$ 9/10 arm0, 8/10 arm1-head). The residual frontier everywhere is the $k=2/k=3$ boundary (Fig.~\ref{fig:perk}).

\subsection{Transfer (v1x, $n=90$, independent factory)}

run11-lineage 61.1 $\to$ \textbf{A-base 76.7 [+67.8, +84.4]} ($\Delta$ vs prior +0.156, $P=.011$) $\to$ structural arms 71.1--72.2 with \textbf{no significant transfer cost} (arm2$-$A-base $=-0.044$ [$-0.133$, $+0.044$], $P=.871$). Clean balanced data is itself a strong transfer vehicle; the readout interventions buy in-distribution gains at no measured cross-factory cost.

\begin{figure}[t]
\centering
\includegraphics[width=\columnwidth]{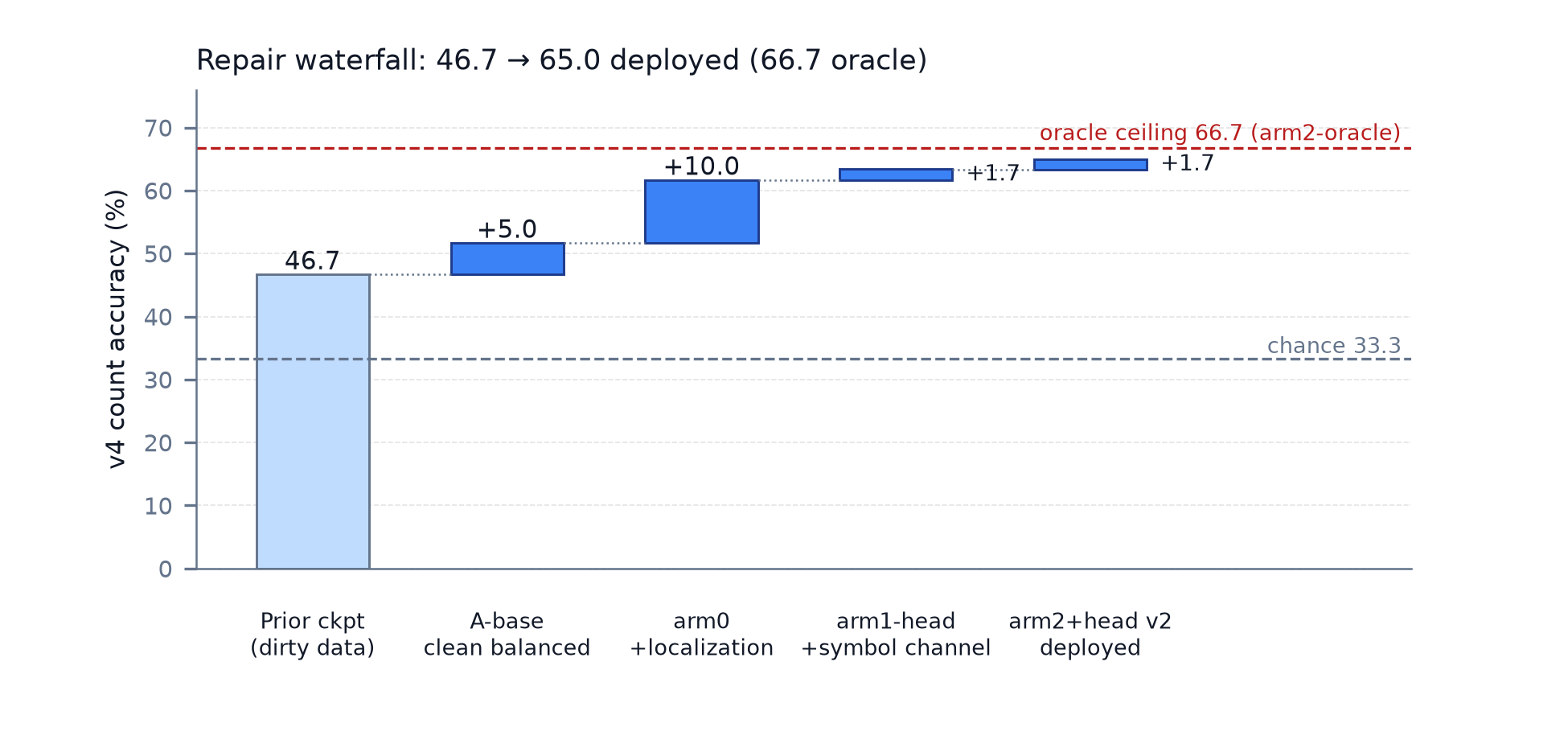}
\caption{\textbf{Repair waterfall.} From the dirty-data checkpoint (46.7) to the deployed full stack (65.0): clean balanced data (+5.0), episode-localization supervision (+10.0, $P=.047$), the event-symbol side channel (+1.7 on top of arm0; +11.7 vs A-base, $P=.008$), and the edge-band head v2 (+1.7). The oracle ceiling (66.7) sits 1.7 points above deployment.}
\label{fig:waterfall}
\end{figure}

\section{Real-Video Validation (REALCOUNT)}
\label{sec:real}

Everything to this point lives in one synthetic paradigm: a red square crossing real-video backgrounds. We therefore built \textbf{REALCOUNT}, a real-video exit-and-reappear counting set, and re-ran the diagnosis and both readout repairs on natural footage.

\textbf{Data and audit.} REALCOUNT draws on VOT2018-LT and VOT2019-LT (85 long-term tracking sequences, selected \emph{because} the target leaves the frame). Absence is encoded in the tracking ground truth itself (the box becomes NaN while the target is out of view---verified at frame level); episodes are visible runs of $\ge$1.0 s separated by $\ge$0.4 s absences, and $k$ is the number of re-entries in a 5--25 s window. Candidate windows ($k\in\{1,2,3\}$) were sampled around re-entries, then every candidate passed a two-layer audit: 1-fps contact-sheet inspection of all 60 initially selected windows, followed by three independent blind review passes that frame-verified every entry/exit boundary with the ground truth hidden. \textbf{Only 18 of 78 audited windows (23\%) survived}---rejected for multi-instance scenes (15), first-person/onboard cameras (5), occlusion-vs-exit ambiguity (3), GT-vs-visual mismatches (7), class mismatches (7), unverifiable targets (6), and query ambiguity (2). The final set is $n=18$ ($k\in\{1,2,3\}$: 5/7/6; 11 sequences; 5--25 s), every label blind-audited; the audit protocol and full rejection taxonomy are released with the set.

\begin{table}[t]
\centering
\footnotesize
\caption{Real-video validation. $n=18$; every label verified by blind frame-level review; paired deltas vs the arm2 reader over $B=10{,}000$ resamples.}
\begin{tabular}{@{}L{0.62\columnwidth}L{0.32\columnwidth}@{}}
\toprule
Arm & REALCOUNT acc (chance 33.3\%) \\
\midrule
run11 reader, no symbols & 50.0\% \\
arm2 reader, no symbols & \textbf{55.6\%} \\
arm2 + oracle GT timelines & 44.4\% ($P(\Delta\le0)=1.0$) \\
arm2 + REALCOUNT-trained head & 33.3\% ($P(\Delta\le0)=1.0$) \\
30B zero-shot & 44.4\% ($k{=}1$: 0/5; $k{=}3$: 5/6) \\
P1 slot-presence linear probe & 72.6\% $\approx$ majority 71.4\% \\
P2 pooled-count linear probe & 33.3\% = duration = chance \\
REALCOUNT-trained detector & 58.0\% $<$ 71\% (all-present) \\
\bottomrule
\end{tabular}
\label{tab:real}
\end{table}

\textbf{Findings.} (i) \emph{The boundary persists.} No unsigned system is reliably above chance under a two-sided test (best 55.6\%, $n=18$; two-sided binomial $p=.076$, one-sided $p=.043$); natural footage is not easier than the synthetic paradigm. (ii) \emph{The diagnosis sharpens.} On real video, even local presence is not linearly readable from the connector's tokens (72.6\% vs 71.4\% majority; synthetic: 93.3\% vs 74\%) and global count decoding sits at chance---``present but unread'' becomes ``presence itself linearly unread''. (iii) \emph{The symbol channel does not transfer zero-shot.} Injecting \textbf{oracle ground-truth timelines hurts} (harmful in every bootstrap resample), and the REALCOUNT-trained detector transfers poorly (slot accuracy 58.0\% vs the 71\% trivial all-present bound; absent-class recall 19\%); the timeline length distribution (up to 50 slots vs $\le$16 at training) compounds the mismatch. The deployability claim of Sec.~\ref{sec:arm2} is therefore bounded to the synthetic paradigm unless the channel is trained in-domain; the in-domain result separates the oracle channel from the deployed head (Sec.~\ref{sec:ceiling}). (iv) \emph{What does transfer} is the readout supervision itself: the arm2 reader remains the strongest counter without symbols (55.6\%). Closing the loop on real video requires in-domain training, and our first attempts delimit the difficulty. Naive full-layer SFT on the REALCOUNT training windows (warm-started from arm2, halved learning rate, one epoch) collapses the reader's output distribution on held-out clips (55.6\% $\to$ unparseable); a pipeline audit further shows that 56 of 131 windows exceeded our conservative 8,000-token training budget (the backend fails above seq $\approx$8,757, where 28 heads $\times$ seq$^2>2^{31}$) and were silently skipped, so the effective SFT set was $\approx$75 windows. Parameter-efficient adaptation (rank-16 LoRA \citep{hu2022lora} on all language-attention projections, connector frozen) eliminates the collapse---18/18 answers parseable and synthetic accuracy preserved (53.3\% $\to$ 56.7\%)---but does not restore real-video counting (27.8\%; 50.0\% without synthetic replay, vs.\ 55.6\% baseline). The residual failure is an \emph{answer-prior collapse}: both PEFT variants answer ``three'' for most clips ($k{=}2$: 0/7, $k{=}3$: 5/6) and never recover the two-event cell, even in-sample (61.3\% overall, but $k{=}1$ 1/11 and $k{=}2$ 16/33). Doubling the real training set (75 $\to$ 145 windows, with better $k$ balance) leaves held-out accuracy unchanged---9/18 in both runs ($\Delta=0$, $P=0.63$)---while merely redistributing the prior: the $k{=}2$ cell recovers from 0/7 to 5/7, the $k{=}3$ cell falls from 5/6 to 2/6, and 12 of 18 answers become the training-set mode (``two''). Data quantity in this regime moves the marginal answer distribution, not the reading skill. We read this as evidence that the bottleneck is the evidence and the paired supervision that teaches its use, not the optimizer. Sec.~\ref{sec:ceiling} tests the remaining escape route---paired symbol training---and a 3-epoch schedule shows that the 1-epoch null is budget-limited: with oracle timelines the reader reaches 94.4\% sequence-disjoint (100\% strict on the full set), while the deployed 2-epoch head reaches 61.1\% and does not beat a shuffled control (61.1\% in seed 42, 66.7\% in seed 43).

\subsection{The adaptation ceiling}
\label{sec:ceiling}

Table~\ref{tab:ceiling} collects five in-domain adaptation levers evaluated on the same 18-clip benchmark at a 1-epoch budget, each warm-started from the arm2 reader and trained only on REALCOUNT windows outside the evaluation spans. We additionally test a sixth---paired symbol training, the missing arm that the synthetic result calls for---at the same 1-epoch budget in Table~\ref{tab:paired}, and then extend its oracle-timeline variant to three epochs in Table~\ref{tab:curve}.

\begin{table}[t]
\centering
\footnotesize
\caption{In-domain adaptation levers on REALCOUNT ($n=18$; chance 33.3\%). v4 = synthetic retention ($n=60$). All arms warm-start from the arm2 reader and train only on REALCOUNT windows outside the evaluation spans. The six levers predate the prompt-class correction; the clean retrain covers the symbol-channel arms (Tables~\ref{tab:paired}--\ref{tab:curve}).}
\begin{tabular}{@{}L{0.29\columnwidth}L{0.13\columnwidth}L{0.10\columnwidth}L{0.33\columnwidth}@{}}
\toprule
Lever (trainable) & REAL-COUNT & v4 & Observation \\
\midrule
arm2 baseline (---) & \textbf{55.6\%} & 53.3\% & reference \\
Readout LoRA + replay (10.1M) & 27.8\% & 56.7\% & format fixed; prior collapse \\
Readout LoRA, no replay & 50.0\% & 51.7\% & parity; prior stays \\
Readout LoRA, 145 win. & 50.0\% & 58.3\% & prior $\to$ mode \\
Connector projection (11.6M) & 44.4\% & 50.0\% & no gain; no prior shift \\
Localization objective & 44.4\% & \textbf{61.7\%} & boundary recall 18.9\% \\
Encoder last-2 blocks (25.8M) & 44.4\% & 51.7\% & weights move; probe unchanged \\
\bottomrule
\end{tabular}
\label{tab:ceiling}
\end{table}

\textbf{At the 1-epoch budget, no deployable lever exceeds the unadapted reader} (best deployable 50.0\%, non-significant deltas; paired symbol training included; the GT-timeline head reader reaches 61.1\%, but it is not deployable); synthetic retention for the six levers ranges 50--63\% without symbols (65--66.7\% with), and the clean 1-epoch oracle retains 65.0\%. The localization arm is the most diagnostic: it produces parseable JSON for all 18 clips and matches $k$ in 10, yet boundary recall at IoU$\ge$0.5 is only 18.9\%; 12 of 14 long ($>$3 s) ground-truth episodes are fragmented into two or more predicted intervals, and predicted durations have p90 2.7 s against 7.5 s, despite 25 training episodes longer than 6 s. What this arm learns is a duration/count prior, not boundary reading. A linear probe on frozen six-band VideoMAE features reaches boundary AUC \textbf{0.62} (permutation control 0.51; edge recall 0.33, absent recall 0.21), and a nonlinear head on the same features does no better on edges (0.25)---so readout capacity is not the limit. Fine-tuning the last two VideoMAE blocks changes weights (max$|\Delta|\approx7\times10^{-4}$) but leaves both accuracy and the probe unchanged (AUC 0.766/recall 0.636 before and after); the same probe reaches 0.766 within clips but only 0.62 across windows, i.e., the boundary signal exists in-distribution but does not transfer. At this data scale (75--145 windows) the 1-epoch bench fails on weak boundary evidence and poor cross-clip transfer. The symbol-channel arm is the exception once the budget is increased (Table~\ref{tab:curve}): a 3-epoch schedule installs counting with oracle timelines (94.4\% sequence-disjoint; 100\% strict on all 145 windows), while the deployed 2-epoch head reaches 61.1\% and does not beat a shuffled-symbol control (61.1\% in seed 42, 66.7\% in seed 43), so the remaining gap is detector fidelity and reader--detector mismatch rather than adapter capacity. A controlled timeline-corruption analysis (Appendix~\ref{app:curve}, Table~\ref{tab:timeline}) sharpens this: absent-slot loss dominates $k{=}1$, transition loss dominates $k{=}3$, and repairing both with oracle patches recovers 17/18.

\begin{table}[t]
\centering
\footnotesize
\caption{Paired symbol training on REALCOUNT at the 1-epoch budget ($n=18$; chance 33.3\%). Rows: timelines used during training (145 windows, same LoRA recipe); columns: timelines at test. The unadapted arm2 without symbols scores 55.6\%.}
\begin{tabular}{@{}lcc@{}}
\toprule
Training timelines & head @ test & GT @ test \\
\midrule
none (arm2, zero-shot) & 44.4\% & 50.0\% \\
deployed head (145 win.) & 50.0\% & 61.1\% \\
oracle GT (145 win.) & 44.4\% & 44.4\% \\
\bottomrule
\end{tabular}
\label{tab:paired}
\end{table}

\textbf{Paired training at the 1-epoch budget does not install deployable symbol reading.} No deployable cell in Table~\ref{tab:paired} exceeds the unadapted reader: the head-trained reader reaches 50.0\% on head timelines (vs.\ 44.4\% for the same channel zero-shot, $\Delta=+1$ clip, $P=0.41$), while its GT-timeline cell reaches 61.1\%---above the 55.6\% baseline, but not deployable. The GT-trained reader is at 44.4\% under both timeline sources, i.e., it does not install symbol reading at this budget. The clean 1-epoch oracle retains 65.0\% on v4, so this is not catastrophic forgetting. The null is budget-specific, not a capacity limit: extending the oracle-timeline arm to 3 epochs reverses it (Table~\ref{tab:curve}).

\textbf{The 1-epoch null is a budget artifact.} Extending the oracle-timeline arm to three epochs reverses the 1-epoch null. On the episode- and prompt-corrected manifest, a 3-epoch \emph{sequence-disjoint} run (117 windows from the 19 sequences that never appear in evaluation) reaches \textbf{94.4\%} (17/18; $k{=}1$ 5/5, $k{=}2$ 7/7, $k{=}3$ 5/6; v4 85.0\%). Training on all 145 corrected windows reaches \textbf{100\% strict} in both seeds tested (18/18; v4 73.3\%/85.0\%); because the full set includes 28 windows from sequences that overlap the evaluation set, we treat the sequence-disjoint 94.4\% as the primary generalization result and report the full-set 100\% as a secondary strict result. Two controls bound the mechanism: a 3-epoch no-symbol run reaches 50.0\% and a content-free shuffled-symbol run 61.1\% in seed 42 (66.7\% in seed 43), so the oracle gain requires informative symbols. The deployed 2-epoch head reaches 61.1\% (11/18) in all three seeds (42--44), no better than the shuffled control (61.1\% seed 42, 66.7\% seed 43) and one clip above the unadapted reader (55.6\%); the same reader reaches 72.2--88.9\% on GT timelines (seed 42: 77.8\%, 14/18), and the sequence-disjoint oracle reaches 94.4\% in both seeds tested (42--43; v4 75--85\%). The sequence-disjoint oracle reader reaches 66.7\% (12/18) on head timelines. The residual 33.3-point gap (94.4 vs.\ 61.1) is therefore detector evidence and a head--reader mismatch, not reader capacity: the reader can read clean timelines, but the head's timelines carry too little usable evidence. A head-trained reader does not beat the GT-trained reader on head timelines (61.1 vs.\ 66.7), so noise-matched training does not close the gap at this scale (Table~\ref{tab:curve}).

\section{Discussion}
\label{sec:disc}

\textbf{Why symbols work.} The same models are near-perfect at counting \emph{discrete heterogeneous} events (flashes of distinct colors: 93--100\%). The symbol channel converts continuous same-object evidence into exactly that regime. The division of labor is measurable: the detector supplies structure (84\% edge recall after the band-pooling fix), the LLM supplies symbol manipulation (robust to clustered and missing symbols---it spontaneously learned ``a run of transition markers is one event'', outperforming our hand-written counting rule on the same noisy timelines).

The failure mode we observe is a real-video instance of the preferred-number collapse RepProbe names: answers concentrate on a few numbers \citep{khoshnoodi2026reprobe}. Our 1-epoch readers show the same collapse ($k{=}2$: 0/7; the training-set mode becomes the answer), and the symbol channel removes it by moving aggregation outside the reader: the model counts discrete symbols rather than regressing a count from continuous visual tokens. This is consistent with the abstraction-level account of specialist cues in crowd counting \citep{crowdcue2026}, and it explains why merely marking items is not enough \citep{givethemodelafinger2026}: the symbols must carry event structure and the reader must be trained to consume them.

\textbf{Deployment shape.} The channel costs one token per half-second, requires no base-model retraining to attach (arm0-trained and arm1-trained readers both exploit it), and its detector trains in minutes on frozen features. On the synthetic paradigm the remaining gap is small: the deployed head sits 1.7 points below the oracle, and better detectors translate directly into the ceiling. On REALCOUNT the gap is much larger: the oracle reader reaches 94.4\% sequence-disjoint (and 100\% strict on the full set), but the deployed-timeline reader reaches 61.1\% and the same reader reaches 72.2--88.9\% on GT timelines. The 33.3-point gap is therefore detector evidence, not reader capacity: the reader can read clean timelines, but the head's timelines carry too little usable evidence. A head-trained reader does not beat the GT-trained reader on head timelines (61.1 vs.\ 66.7), and the deployed head does not exceed a content-free shuffled control (61.1\% in seed 42, 66.7\% in seed 43). Noise-matched training alone does not close this gap; better detectors (unfrozen backbones, richer temporal context) and a calibrated head--reader pair are the next step. The channel is not yet deployable on real video; the oracle result establishes that the readout failure is repairable in principle.

\textbf{Limitations.} (i) Synthetic testbeds---necessary for construction-grounded episode labels at this scale; the transfer factory and the REALCOUNT audit (Sec.~\ref{sec:real}) mitigate but do not eliminate domain concerns. (ii) One base family at intervention scale (7B); the 30B replication is prepared but was not run (hardware budget). (iii) $n=60/90$ with reported CIs; per-$k$ cells are smaller still. The main synthetic arms are single-seed (seed 42); on REALCOUNT, the deployed head was repeated across seeds 42--44 (61.1\% each), the sequence-disjoint oracle across 42--43 (94.4\% each), the shuffled control across 42--43 (61.1/66.7\%), and the GT-timeline reader varies 72.2--88.9\%; the remaining arms are single-seed. (iv) The oracle-vs-deployment gap, though small on the synthetic paradigm, is large on REALCOUNT and is set by detector fidelity and by the reader's match to that detector's noise. (v) REALCOUNT is $n=18$: small by construction---every item is blind-audited, and the 23\% audit yield is itself a finding about real-video episode counting---but CIs are wide ($\pm$20pp level) and cross-domain deltas should be read as directional. (vi) The adaptation bench (Sec.~\ref{sec:ceiling}) is negative at the 1-epoch budget; a 3-epoch schedule reverses the symbol-channel arm with oracle timelines (94.4\% sequence-disjoint; 100\% strict on the full set), but the deployed head stays at 61.1\%, no better than a shuffled-symbol control (61.1\% in seed 42, 66.7\% in seed 43) and only one clip above the unadapted reader. The head-trained reader does not beat the GT-trained reader on head timelines (61.1 vs.\ 66.7); the readout failure is repairable with informative timelines, but the deployed head is not yet sufficient. The full-145 100\% is secondary because 28 of its windows overlap the evaluation sequences. (vii) Concurrent counting benchmarks are one to three orders of magnitude larger than REALCOUNT (e.g., 42K clips in RepProbe \citep{khoshnoodi2026reprobe}); we position REALCOUNT as an audited repair testbed, not a coverage benchmark.

\section*{Impact Statement}
This work is diagnostic and benchmark-oriented. On the positive side, it gives the community a measurement protocol for a failure mode that is easy to miss (label priors, leakage, and readout bottlenecks), a blind-audited real-video counting set, and evidence that a 1-epoch in-domain adaptation bench does not fix the failure while oracle event timelines do (94.4\% sequence-disjoint; 100\% strict on the full set), whereas the deployed head remains at 61.1\% in all three seeds and does not beat a shuffled control (61.1\% in seed 42, 66.7\% in seed 43)---which should redirect effort toward detector fidelity and calibrated head--reader pairs rather than adapter tuning. On the negative side, counting and temporal-grounding technology can support surveillance or automated monitoring; we release no person-tracking model, no raw person footage, and no deployment-ready detector. The REALCOUNT clips derive from VOT-LT research sequences under their original licenses; we release audit sheets, manifests, and generation scripts rather than redistributing raw frames. Compute was modest (a single 64\,GB workstation; the 30B model was used for inference only), so the direct environmental cost is small. We see no other specific ethical risks beyond those already present in video-understanding research.

\bibliography{refs}
\bibliographystyle{icml2026}

\appendix

\section{Data-Integrity Findings}
\label{app:integrity}
\begin{enumerate}
\item \textbf{Set-inclusion leakage.} A hard-cut evaluation set was fully contained (20/20) in the training core from an early fine-tune onward, producing a lineage of ``record'' timestamp precisions (11--50 ms). Detection: file-level ID intersection audits as a release gate.
\item \textbf{Memorization masquerading as precision.} The leaked precision survived \emph{time-reversed inputs} unchanged---exceeding every input channel's physical resolution. Rule: \textbf{impossible precision is a leak signal}; input-scrambling (reversal, content freezing) is a cheap probe.
\item \textbf{Formula-reuse label error.} A crossing expression reused across two factories was linear in one (object exits and never returns), silently invalidating $\sim$465 ``continuous crossing'' samples. Rule: \textbf{construction-grounded labels still require independent measurement} (here: presence-episode counting for gap data, centroid-wraparound counting for nogap).
\item \textbf{Training-window episode truncation.} In 64 of the 145 REALCOUNT training windows, every episode end was written to the clip end because the candidate builder used the wrong map key; the resulting episodes overlapped and marked post-exit frames as present. The evaluation set was unaffected. Recomputing episodes from GT removed all overlaps and preserved $k$ for all 145 windows; the oracle-timeline readers in Sec.~\ref{sec:ceiling} and Appendix~\ref{app:curve} were retrained on the episode-repaired manifest. Detection: compare each episode end with the next absence run, never with the clip end.
\item \textbf{Training-prompt class mismatch (closed in this version).} A blind audit of the 145 REALCOUNT training windows found 29 prompts whose object class did not match the tracked target: 27 car windows were prompted as a rider and 2 dog windows as a dragon. Episodes and symbol timelines refer to the tracked target and are unaffected; only the reader's class cue is wrong. The evaluation set is unaffected (class mismatches were rejected during its audit). A corrected-prompt retrain of the core arms is reported in Tables~\ref{tab:ceiling}--\ref{tab:curve}; it closes the disclosure. Detection: compare the prompt class with the tracker's target class for every training window, not only the evaluation set.
\end{enumerate}
All five were found \emph{because} the diagnosis demanded controlled measurement; we expect comparable rates elsewhere in synthetic video benchmarks.

\section{Reproducibility and Disclosure}
\label{app:repro}
Every number in this paper is reproduced by the anonymized supplementary package (README, the exact per-run argument dictionary extracted from each checkpoint, and one JSONL row per evaluation sample): \texttt{verify\_results.py} re-derives all table values from the archived records without model weights or videos. The archival version will receive a permanent DOI upon acceptance.

\textbf{Compute.} All 7B training and evaluation ran on a single Apple M5 Pro workstation with 64 GB unified memory (MPS backend); no GPU cluster was used. Each 2,159-step arm takes 8--12 h, each 2,939-step arm about 12 h, a 435-step run about 4 h, and the presence heads train in minutes on cached features. The 30B baseline is 4-bit inference only; its fine-tuning replication was not run.

\textbf{AI-assistance disclosure.} The research programme---problem selection, experimental design and its approval, hardware and budget decisions, stopping rules, and acceptance of reported results---was directed by the human author. All code, data factories, experiments, and manuscript text were produced by an AI coding agent operating under that direction. No AI system is an author of this work: responsibility and accountability for every claim rest entirely with the human author, whose obligation is to verify each result against the archived records before submission.

\section{Controls and Cross-Timeline Cells}
\label{app:curve}
\begin{table}[h]
\centering
\footnotesize
\setlength{\tabcolsep}{4pt}
\caption{REALCOUNT controls on the episode- and prompt-corrected training manifest ($n=18$; chance 33.3\%). The full-145 row includes 28 windows from sequences that overlap the evaluation set; the sequence-disjoint row is the primary generalization result. @GT/@head denote the timelines used at test. v4 = synthetic retention ($n=60$). Seed-42 values; seed variance in Table~\ref{tab:seedvar}.}
\begin{tabular}{@{}lcc@{}}
\toprule
Condition & REALCOUNT & v4 \\
\midrule
1-epoch oracle (corrected) & 44.4 & 65.0 \\
3-epoch oracle, full 145 & 100.0 & 73.3 \\
3-epoch oracle, seq-disjoint 117 & \textbf{94.4} & 85.0 \\
3-epoch no symbols & 50.0 & 61.7 \\
3-epoch shuffled symbols & 61.1 & 58.3 \\
2-epoch deployed head @head & 61.1 & 53.3 \\
2-epoch deployed head @GT & 77.8 & --- \\
3-epoch seq-disjoint @head & 66.7 & --- \\
\bottomrule
\end{tabular}

\label{tab:curve}
\end{table}

\textbf{Controls.} On the episode- and prompt-corrected manifest, 1 epoch reaches 44.4\%; 3 epochs reach 100\% strict (18/18) on all 145 windows and \textbf{94.4\%} (17/18) with sequence-disjoint training. The full 145 set includes 28 windows from sequences that overlap the evaluation set, so the sequence-disjoint result is the primary generalization claim. No-symbol (50.0\%) and shuffled-symbol (61.1\% in seed 42, 66.7\% in seed 43) controls show that the oracle gain requires informative symbols, not extra in-domain training or a content-free token stream. Synthetic retention is 65.0\% (1 epoch), 73.3\% (full 3 epochs), 85.0\% (sequence-disjoint, seed 42; 75.0\% in seed 43), 61.7\% (no symbols), 58.3\% (shuffled, seed 42; 63.3\% in seed 43), and 53.3\% (deployed head, seed 42; 60.0/50.0 in seeds 43/44); see Table~\ref{tab:seedvar}. The full-145 and sequence-disjoint arms have overlapping v4 retention (73.3--85.0 vs.\ 75--85); the REALCOUNT difference is one clip (100 vs.\ 94.4). We treat sequence-disjoint as the primary generalization result because 28 of the full-145 windows overlap the evaluation sequences, not because it has better retention.

\textbf{Cross-timeline cells.} The 2-epoch deployed-head reader reaches 61.1\% on head timelines and 77.8\% on GT timelines (seed 42); the sequence-disjoint oracle reader reaches 66.7\% on head timelines (94.4\% on GT). A head-trained reader does not beat the GT-trained reader on head timelines (61.1 vs.\ 66.7), so noise-matched training does not close the gap at this scale. The deployed head's head-timeline score does not exceed the shuffled-symbol control (61.1\% in seed 42, 66.7\% in seed 43); the residual gap is detector evidence, not reader capacity.

\begin{table}[h]
\centering
\footnotesize
\setlength{\tabcolsep}{4pt}
\caption{Seed variance for the REALCOUNT claims ($n=18$; chance 33.3\%; v4 $n=60$). The deployed head's head-timeline score is identical across seeds; the shuffled control and the GT-timeline reader vary; v4 retention varies. The full-145 and sequence-disjoint oracles were not repeated for seed 44.}
\begin{tabular}{@{}lccc@{}}
\toprule
Condition & seed 42 & seed 43 & seed 44 \\
\midrule
deployed head @head & 61.1 & 61.1 & 61.1 \\
deployed head @GT & 77.8 & 72.2 & 88.9 \\
shuffled symbols @GT & 61.1 & 66.7 & --- \\
full oracle @GT & 100.0 & 100.0 & --- \\
v4 full oracle & 73.3 & 85.0 & --- \\
seq oracle @GT & 94.4 & 94.4 & --- \\
v4 seq oracle & 85.0 & 75.0 & --- \\
v4 deployed head & 53.3 & 60.0 & 50.0 \\
v4 shuffled symbols & 58.3 & 63.3 & --- \\
\bottomrule
\end{tabular}
\label{tab:seedvar}
\end{table}

\textbf{Seed variance.} Table~\ref{tab:seedvar} repeats the key claims across seeds. The deployed head is 61.1\% (11/18) on head timelines in all three seeds, so the negative deployment result is not seed noise. The content-free shuffled control is 61.1\% in seed 42 and 66.7\% in seed 43, so the deployed head never exceeds it and in seed 43 sits one clip below it, which strengthens rather than weakens the negative deployment conclusion. The GT-timeline score varies from 72.2\% to 88.9\%, so the reader's ability to use clean timelines is robust in direction but variable in size. The full-145 oracle is 100\% in both seeds tested (v4 73.3/85.0\%); the sequence-disjoint oracle is 94.4\% in both seeds tested; its v4 retention is 75--85\%, and the deployed head's v4 retention is 50--60\%. These ranges should replace single-seed point estimates for the secondary retention claims.

\textbf{What the head loses.} To attribute the 33.3-point gap, we evaluate the same sequence-disjoint reader on controlled timeline variants ($n=18$; all patches copy GT symbols and are oracle interventions, not deployable results). Randomly flipping 10\% of GT slots drops accuracy from 94.4\% to 66.7\%, and flipping 40\% drops it to 38.9\%; the reader is brittle to generic slot noise. The errors are not uniform, however. Filling every absent slot (removing gaps) leaves $k{=}3$ at 6/6 but drops $k{=}1$ from 5/5 to 0/5; removing transitions leaves $k{=}1$ at 5/5 but drops $k{=}3$ from 5/6 to 3/6. The deployed head loses both: 5/18 clips contain no absent slot at all (mean predicted absences 2.5 vs.\ 6.6 in GT; boundary F1 0.38). Repairing only absent slots raises accuracy from 12/18 to 14/18, entirely in $k{=}1$ (3/5 $\to$ 5/5) and without breaking any clip. Repairing only boundary windows leaves the total at 12/18: it fixes two $k{=}3$ clips but breaks two $k{=}1$ clips, because transitions are added while the absence structure is still missing. Repairing both reaches \textbf{17/18 (94.4\%)}, matching the GT-timeline ceiling. The remaining error is a different clip from the one the GT timeline misses, so this is not an exact recovery of GT but a complementary 17/18. These oracle repairs bound what better event evidence could buy; the deployed head remains at 61.1\%.

\begin{table}[h]
\centering
\footnotesize
\setlength{\tabcolsep}{5pt}
\caption{Timeline corruption and repair on REALCOUNT ($n=18$; chance 33.3\%). All rows use the sequence-disjoint oracle reader. ``+'' rows copy GT symbols into the deployed-head timeline and are oracle interventions, not deployable results.}
\begin{tabular}{@{}lccc c@{}}
\toprule
Timeline condition & $k{=}1$ & $k{=}2$ & $k{=}3$ & All \\
\midrule
GT & 5/5 & 7/7 & 5/6 & 17/18 \\
GT, all absent slots filled & 0/5 & 5/7 & 6/6 & 11/18 \\
GT, transitions removed & 5/5 & 6/7 & 3/6 & 14/18 \\
\midrule
Deployed head & 3/5 & 5/7 & 4/6 & 12/18 \\
\quad + GT absent slots & 5/5 & 5/7 & 4/6 & 14/18 \\
\quad + GT boundary windows & 1/5 & 5/7 & 6/6 & 12/18 \\
\quad + both & 5/5 & 6/7 & 6/6 & \textbf{17/18} \\
\bottomrule
\end{tabular}
\label{tab:timeline}
\end{table}

\section{Additional Adaptation Attempts}
\label{app:additional}
Beyond Table~\ref{tab:ceiling} and Table~\ref{tab:curve}, we tested two further detector-side levers on REALCOUNT.

\textbf{Scope.} The detector-side experiments below predate the prompt-corrected retrain; their end-to-end comparisons use the pre-clean 72.2\% head--reader pair and are reported for completeness, not as current deployment estimates. The clean retrain's deployed pair is 61.1\%.

\textbf{Presence-head redesign.} Temporal-difference features (adjacent-slot and absolute differences), $\pm2$-slot context, a small temporal convolutional network, and an absence-heavy expansion (631 windows with $\ge20\%$ absence added to the 1{,}000 training windows) improved edge recall from 25\% to 32--41\%, but absence recall stayed at 21--29\% and the count proxy at 7--11 of 18 clips. With the reader fixed, end-to-end accuracy ranged from 39\% to 78\% across variants and seeds, with no reliable gain over the pre-clean 72.2\% deployed pair, because the injected timeline distribution shifted away from the reader's training distribution.

\textbf{Head-specific encoder adaptation.} We fine-tuned a separate copy of VideoMAE (the reader's connector was left untouched), unfreezing its last 2 or 4 blocks jointly with the head under a two-stage schedule (head frozen for 300 steps, then joint fine-tuning; best checkpoint by held-out slot/edge/count score). This raised slot accuracy from 61.4\% to 62.1--62.9\% and the count proxy to 11 of 18, but the original reader fell to 61--67\% on the shifted timelines. Retraining the reader for two epochs on the $K{=}4$ adapted timelines reaches 72.2\% on REALCOUNT---the same as the pre-clean head--reader pair; the clean retrain's deployed pair is 61.1\%---while synthetic retention drops from 56.7\% to 41.7\%.

\textbf{Reading.} Detector-side changes at this scale either do not add count-carrying content to the timeline or require a co-adapted reader, and the co-adaptation costs synthetic retention. Neither closes the gap to the 94.4\% sequence-disjoint oracle result. This is consistent with the boundary probe and with the view that the remaining bottleneck is the evidence recovered by the frozen visual encoder, not the capacity of the presence head or the adapter.

\section{Artifact Index}
\label{app:artifacts}
{\raggedright
The anonymized package contains the data factories (\texttt{make\_count\_v4.py}, \texttt{make\_count\_v1x.py}, \texttt{make\_realcount\_v1.py}, \texttt{make\_realcount\_train2.py}), training entry points (\texttt{train\_m2.py}; LoRA adapter \texttt{src/lora.py}; paired-mix builder \texttt{make\_peft\_mix.py}), evaluation and probe scripts (\texttt{eval\_motion\_qa.py}, \texttt{eval\_count\_loc.py}, \texttt{eval\_realcount\_loc.py}, \texttt{probe\_realcount*.py}, \texttt{bootstrap\_ci.py}), the detector (\texttt{train\_presence\_head\_v2.py}), the released presence-head weights, a dependency-free verifier (\texttt{verify\_results.py}), and per-run records (the exact argument dictionary plus \texttt{records.jsonl}) for every table and figure; \texttt{paper\_figs.py} regenerates Figs.~\ref{fig:diagnosis}--\ref{fig:perk}. The repository will be anonymized and archived with a permanent DOI upon acceptance.
\par}

\end{document}